\documentclass[letterpaper,10pt,conference]{IEEEtran}
\usepackage{subeqnarray}
\usepackage{amsfonts}
\usepackage{amsmath}
\usepackage{amssymb}
\usepackage{graphicx}
\usepackage{color}
\graphicspath{{images/}}
\usepackage{booktabs}
\usepackage[keeplastbox]{flushend}

\newcommand{\specialcell}[2][c]{%
  \begin{tabular}[#1]{@{}l@{}}#2\end{tabular}}
  
\newcommand{\ra}[1]{\renewcommand{\arraystretch}{#1}}

\title{Offline Handwritten Signature Verification - Literature Review}

\author{Luiz G. Hafemann$^1$, Robert Sabourin$^1$ and Luiz S. Oliveira$^2$ \\
$^1$ \'Ecole de technologie sup\'erieure, Universit\'e du Qu\'ebec, Montreal, Canada\\ 

e-mail: lghafemann@livia.etsmtl.ca, robert.sabourin@etsmtl.ca

\\$^2$ Federal University of Paran\'a, Curitiba, PR, Brazil\\

e-mail: luiz.oliveira@ufpr.br}


\begin{document}
\IEEEoverridecommandlockouts

\IEEEpubid{\makebox[\columnwidth]{978-1-5386-1842-4/17/\$31.00~\copyright{}2017 IEEE\hfill} \hspace{\columnsep}\makebox[\columnwidth]{ }}

\maketitle

\begin{abstract}
The area of Handwritten Signature Verification has been broadly researched in the last decades, but remains an open research problem. The objective of signature verification systems is to discriminate if a given signature is genuine (produced by the claimed individual), or a forgery (produced by an impostor). This has demonstrated to be a challenging task, in particular in the offline (static) scenario, that uses images of scanned signatures, where the dynamic information about the signing process is not available. Many advancements have been proposed in the literature in the last 5-10 years, most notably the application of Deep Learning methods to learn feature representations from signature images. In this paper, we present how the problem has been handled in the past few decades, analyze the recent advancements in the field, and the potential directions for future research.
\end{abstract}


\section{Introduction}

Biometrics technology is used in a wide variety of security applications.
The aim of such systems is to recognize a person based on physiological or behavioral traits. In the first case, the recognition is based on measurements of biological traits, such as the fingerprint, face, iris, etc. The later case is concerned with behavioral traits such as voice and the handwritten signature \cite{jain_introduction_2004}.

Biometric systems are mainly employed in two scenarios: verification and identification. In the first case, a user of the system claims an identity, and provides the biometric sample. The role of the verification system is to check if the user is indeed who he or she claims to be. In the identification case, a user provides a biometric sample, and the objective is to identify it among all users enrolled in the system.

The handwritten signature is a particularly important type of biometric trait, mainly due to its ubiquitous use to verify a person's identity in legal, financial and administrative  areas. One of the reasons for its widespread use is that the process to collect handwritten signatures is non-invasive, and people are familiar with the use of signatures in their daily life \cite{plamondon_online_2000}.

Signature verification systems aim to automatically discriminate if the biometric sample is indeed of a claimed individual. In other words, they are used to  classify query signatures as genuine or forgeries. Forgeries are commonly classified in three types: random, simple and skilled (or simulated) forgeries. In the case of random forgeries, the forger has no information about the user or his signature and uses his own signature instead. In this case, the forgery contains a different semantic meaning than the genuine signatures from the user, presenting a very different overall shape. In the case of simple forgeries, the forger has knowledge of the user's name, but not about the user's signature. In this case, the forgery may present more similarities to the genuine signature, in particular for users that sign with their full name, or part of it. In skilled forgeries, the forger has access for both the user's name and signature, and often practices imitating the user's signature. This result in forgeries that have higher resemblance to the genuine signature, and therefore are harder to detect.

Depending on the acquisition method, signature verification systems are divided in two categories: online (dynamic) and offline (static). In the online case, an acquisition device, such as a digitizing table, is used to acquire the user's signature. The data is collected as a sequence over time, containing the position of the pen, and in some cases including additional information such as the pen inclination, pressure, etc. In offline signature verification, the signature is acquired after the writing process is completed. In this case, the signature is represented as a digital image \cite{impedovo_automatic_2008}. 

Over the last few decades, some key survey papers have summarized the advancements in the field, in the late 80's \cite{plamondon_automatic_1989}, 90's \cite{leclerc_automatic_1994} and 2000's \cite{impedovo_automatic_2008}. Some recent advancements have been consolidated in more recent literature reviews: Impedovo et al. \cite{impedovo_handwritten_2012} provide an update over the authors' previous review \cite{impedovo_automatic_2008}, focusing on advancements such as new acquisition devices (mostly for online signatures) and signature representations; Shah et al. \cite{shah_appraisal_2015} present a critical evaluation of 15 signature verification systems proposed in the literature, classifying each work according to the feature extraction methods, classifiers and overall strengths and limitations of the systems. These reviews, on the other hand, do not capture more recent trends in the field, in particular the usage of Deep Learning methods applied for handwritten signatures. Such methods have demonstrated superior results in multiple benchmarks, and are reviewed in the present work.

This paper is organized as follows: we start by formalizing the problem at hand, and list the main datasets that are available to evaluate such systems. We then describe the techniques used for each process of the pipeline for training a system: Preprocessing, Feature Extraction and model training, and finally we summarize the recent progress and potential areas for future research.

\section{Problem Statement}
\label{sec:problem_statement}

The problem of automatic handwritten signature verification is commonly modeled as a verification task: given a learning set $\mathcal{L}$, that contains genuine signatures from a set of users, a model is trained. This model is then used for verification: a user claims an identity and provides a query signature $X_\text{new}$. The model is used to classify the signature as genuine (belonging to the claimed individual) or forgery (created by someone else). To evaluate the performance of the system, we consider a test set $\mathcal{T}$, consisting of genuine signatures and forgeries. The signatures are acquired in an \emph{enrollment} phase, while the second phase is referred to \emph{operations} (or \emph{classification}) phase. 

If a single model is used to classify images from any user, we refer to it as a \emph{writer-independent} (WI) system. If one model is trained for each user, it is referred as a \emph{writer-dependent} (WD) system. 
For WI systems, the common practice is to train and test the system with a different subset of users. In this case, we consider a development set $\mathcal{D}$ (which is used to train the WI model), and an exploitation set $\mathcal{E}$, which represent the users enrolled to the system (and is further divided in $\mathcal{L}$ and $\mathcal{T}$, as indicated above).

Most work in the literature do not use skilled forgeries for training (e.g. \cite{rivard_multi-feature_2013}, \cite{eskander_hybrid_2013}). Other work use skilled forgeries for training writer-independent classifiers, testing these classifiers in a separate set of users (e.g. \cite{yilmaz_score_2016}, \cite{rantzsch_signature_2016}, \cite{hafemann_pr_2017}); lastly, some papers use skilled forgeries for training writer-dependent classifiers, and test these classifiers in a separate set of genuine signatures and forgeries from the same set of users. We restrict our analysis to methods that do not rely on skilled forgeries for the users enrolled in the system (the set $\mathcal{E}$), since this is not the case in practical applications. We do consider, however, that a dataset consisting of genuine signatures and forgeries is available for training writer-independent classifiers (the set $\mathcal{D})$, where the users from this dataset are not used for evaluating the performance of the classifier. This is reasonable for a practical application, since it is possible for an institution to collect forgeries for \emph{some} users (e.g. by detecting actual forgery attempts), that could be used for training WI systems.


\subsection{Challenges}

\begin{figure}
  \centering
  \includegraphics[scale=1]{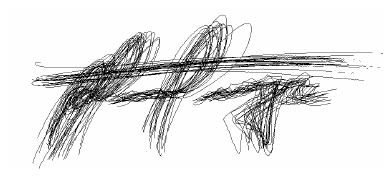}
  \caption{Superimposed examples of multiple signatures of the same user. We can notice a high intra-class variability of the signatures of the user \cite{justino_off-line_2000}.}
  \label{fig:variability}
\end{figure}

One of the main challenges for the signature verification task is having a high intra-class variability. Compared to physical biometric traits, such as fingerprint or iris, handwritten signatures from the same user often show a large variability between samples. This problem is illustrated in Figure \ref{fig:variability}. This issue is aggravated with the presence of low inter-class variability when we consider skilled forgeries. These forgeries are made targeting a particular individual, where a person often practices imitating the user's signature. For this reason, skilled forgeries tend to resemble genuine signatures to a great extent.

Another important challenge for training an automated signature verification system is the presence of partial knowledge during training. In a realistic scenario, during training we only have access to genuine signatures for the users enrolled to the system. During operations, however, we want the system not only to be able to accept genuine signatures, but also to reject forgeries. This is a challenging task, since during training a classifier has no information to learn what exactly distinguishes a genuine signature and a forgery for the users enrolled in the system.

Lastly, the amount of data available for each user is often very limited in real applications. During the enrollment phase, users are often required to supply only a few samples of their signatures. In other words, even if there is a large number of users enrolled to the system, a classifier needs to perform well for a new user, for whom only a small set of samples are available.

\section{Datasets}

A large amount of research in automated signature verification has been conducted with private datasets. This makes it difficult to compare relate work, since an improvement in classification performance may be attributed to a better method, or simply to a cleaner or simpler database. In the last decade, however, a few signature datasets were made available publicly for the research community, addressing this gap.

The process to acquire the signature images follows similar steps for most of the public datasets. The genuine signatures are collected in one or more sessions, and require the user to provide several samples of their signatures. The user receives a form containing many cells, and provide a sample of his/her signature in each cell. The cells often have sizes to match common scenarios such as bank cheques and credit card vouchers \cite{vargas_off-line_2007}. The collection of forgeries follows a different process: the users receive samples from genuine signatures and are asked to imitate the signature one or more times. It is worth noting that the users that provide the forgeries are not experts in producing forgeries. After the forms are collected, they are scanned (often at 300 dpi or 600 dpi), and pre-processed. 

Table \ref{table:datasets} presents a summary of the most commonly used signature datasets. 


\begin{table} \centering
\caption{Commonly used signature datasets}
\ra{1.3}
\resizebox{\columnwidth}{!}{%
\begin{tabular}{@{}lccc@{}} \toprule
Dataset Name & Users & \specialcell{Genuine \\signatures} & Forgeries \\ \midrule

CEDAR \cite{kalera_offline_2004} & 55 & 24 & 24 \\	
MCYT-75 \cite{fierrez-aguilar_off-line_2004}  &	75	& 15 & 15 \\
GPDS Signature 160 \cite{ferrer_offline_2005} &  160 & 24 & 30 \\
\specialcell{GPDS Signature 960 \\~~Grayscale \cite{vargas_off-line_2007}} & 881 & 24 & 30 \\
\specialcell{GPDS Synthetic\\~~Signature\cite{ferrer_synthetic_2013}} & 4000 & 24 & 30 \\
Brazilian (PUC-PR) \cite{freitas_bases_2000} & 60 + 108 & 40 & 10 simple, 10 skilled\footnotemark \\ 	

\bottomrule
\end{tabular}}
\label{table:datasets}
\end{table}

\footnotetext{for 60 users only}

%

\section{Preprocessing}

As with most pattern recognition problems, preprocessing plays an important role in signature verification. Signature images may present variations in terms of pen thickness, scale, rotation, etc., even among authentic signatures of a person. Bellow we summarize the main preprocessing techniques: 
\begin{itemize}

\item \textbf{Signature extraction} - This is an initial step that consists in finding and extracting a signature from a document. This is a particular challenging problem in bank cheques, where the signature is often written on top of a complex background \cite{dimauro_multi-expert_1997}, \cite{djeziri_extraction_1998}. This step is, however, not considered in most signature verification studies, that already consider signatures extracted from the documents.

\item \textbf{Noise Removal} - Scanned signature images often contain noise. A common strategy to address this problem is to apply a noise removal filter to the image, such a median filter \cite{huang_off-line_1997}.  It is also common to apply morphological operations to fill small holes and remove small regions of connected components  \cite{huang_off-line_1997} \cite{yilmaz_score_2016}.

\item \textbf{Size normalization and centering} - Depending on the properties of the features to be used, different size normalization strategies are adopted. The simplest strategy is to crop the signature images to have a tight box on the signature \cite{ghandali_method_2008}.  Another strategy is to user a narrower bounding box, such as cutting strokes that are far from the image centroid, that are often subject to more variance in a user's signature \cite{yilmaz_score_2016}. Other authors use a fixed frame size (width and height), and center the signature in this frame \cite{pourshahabi_offline_2009}, \cite{hafemann_pr_2017}. 

\item \textbf{Signature representation} - Besides just using the gray-level image as input to the feature extractors, other representations have been considered. For instance, using the signature's skeleton, outline, ink distribution, high pressure regions and directional frontiers \cite{huang_off-line_1997}. 


\item \textbf{Signature Alignment} - alignment is a common strategy in online signature verification, but not broadly applied for the offline scenario. Yilmaz \cite{yilmaz_score_2016} propose aligning the signatures for training, by applying rotation, scaling and translation. 
Kalera et al. \cite{kalera_offline_2004} propose a method to perform Rotation normalization, using first and second order moments of the signature image.

\end{itemize}

\section{Feature Extraction}

Offline signature verification has been studied from many perspectives, yielding multiple alternatives for feature extraction. Broadly speaking, the feature extraction techniques can be classified as \emph{Static} or \emph{Pseudo-dynamic}, where pseudo-dynamic features attempt to recover dynamic information from the signature execution process (such as speed, pressure, etc.). Another broad categorization of the feature extraction methods is between \emph{Global} and \emph{Local} features. Global features describe the signature images as a whole - for example, features such as height, width of the signature, or in general feature extractors that are applied to the entire signature image. In contrast, local features describe parts of the images, either by segmenting the image (e.g. according to connected components) or most commonly by the dividing the image in a grid (of Cartesian or polar coordinates), and applying feature extractors in each part of the image.

Recent studies approach the problem from a representation learning perspective \cite{hafemann_ijcnn_2016}, \cite{hafemann_pr_2017}, \cite{rantzsch_signature_2016}, \cite{zhang_multi-phase_2016}: instead of designing feature extractors for the task, these methods rely on learning feature representations directly from signature images. 

\subsection{Handcrafted feature extractors}

A large part of the research efforts on the field has been devoted to finding good feature representations for offline signatures. In this section we summarize the main descriptors proposed for the problem.

\subsubsection{Geometric Features}

Geometric features measure the overall shape of a signature. This includes basic descriptors, such as the signature height, width, caliber (height-to-width ration) and area. More complex descriptors include the count of endpoints and closed loops \cite{baltzakis_new_2001}.
Besides using global descriptors, several authors also generate local geometric features by dividing the signature in a grid and calculating features from each cell. For example, using the pixel density within grids \cite{baltzakis_new_2001}, \cite{el-yacoubi_off-line_2000}, \cite{justino_off-line_2000}.

\subsubsection{Graphometric features}

Forensic document examiners use the concepts of graphology and graphometry to examine handwriting for several purposes, including detecting authenticity and forgery. Oliveira et al. \cite{oliveira_graphology_2005} investigated applying such features for automated signature verification. They selected a subset of graphometric features that could be described algorithmically, and proposed a set of feature descriptors. They considered the following static features: \emph{Calibre} - the ratio of Height / Width of the image; \emph{Proportion}, referring to the symmetry of the signature, \emph{Alignment to baseline} - describing the angular displacement to an horizontal baseline, and \emph{Spacing} - describing empty spaces between strokes.

\subsubsection{Directional features}

Directional features seek to describe the image in terms of the direction of the strokes in the signature. Sabourin \cite{sabourin_off-line_1992} and Drouhard \cite{drouhard_neural_1996} extracted Directional-PDF (Probability Density Function) from the gradient of the signature outline. Rivard et al. \cite{rivard_multi-feature_2013} used this method of feature extraction using grids of multiple scales. Zhang et al. have investigate the usage of pyramid histogram of oriented gradients (PHOG) \cite{zhang_off-line_2010}. This descriptor represents local shapes in a image by a histogram of edge orientations, also in multiple scales.

\subsubsection{Mathematical transformations}

Researchers have used a variety of mathematical transformations as feature extractors. Nemcek and Lin \cite{nemcek_experimental_1974} investigated the usage of a fast \emph{Hadamart} transform and spectrum analysis for feature extraction.  Pourshahabi et al. \cite{pourshahabi_offline_2009} used a \emph{Contourlet} transform as feature extraction, stating that it is an appropriate tool for capturing smooth contours. Coetzer et al. \cite{coetzer_off-line_2005} used the discrete \emph{Radon} transform to extract sequences of observations, for a subsequent HMM training. Deng et al \cite{deng_wavelet-based_1999} proposed a signature verification system based on the \emph{Wavelet} transform. Zouari et al \cite{zouari_identification_2014} has investigate the usage of the \emph{Fractal} transform for the problem.

\subsubsection{Shadow-code}

Sabourin et al. \cite{sabourin_off-line_1992}, \cite{sabourin_extended-shadow-code_1994} proposed an Extended Shadow Code for signature verification. A grid is overlaid on top of the signature image, containing horizontal, vertical and diagonal bars, each bar containing a fixed number of bins. Each pixel of the signature image is then projected to its closest bar in each direction, activating the respective bin. The count of active bins in the projections is then used as a descriptor of the signature. This feature extractor has been used by Rivard \cite{rivard_multi-feature_2013} and Eskander \cite{eskander_hybrid_2013} with multiple resolutions, together with directional features, to achieve promising results on writer-independent and writer-dependent classification, respectively.

\subsubsection{Texture features}

Texture features, in particular variants of Local Binary Patterns (LBP), have been used in many experiments in recent years. The LBP operator describe the local patterns in the image, and the histogram of these patterns is used as a feature descriptor. LBP variantions have  been used in many studies \cite{yilmaz_offline_2011}, \cite{yilmaz_score_2016}, 
\cite{serdouk_combination_2014}, \cite{serdouk_orthogonal_2014}, \cite{hu_offline_2013}, and have demonstrated to be among the best hand-crafted feature extractors for this task. Another important texture descriptor is GLCM (Gray Level Co-occurrence Matrix). This feature uses relative frequencies of neighboring pixels, and was used in a few papers \cite{hu_offline_2013}, \cite{vargas_off-line_2011}.

\subsubsection{Interest point matching}

Interest point matching methods, such as SIFT (Scale-Invariant Feature Transform) and SURF (Speeded Up Robust Features) have been largely used for computer vision tasks.
Ruiz-del-Solar et al. \cite{ruiz-del-solar_offline_2008} used SIFT to extract local interest points from both query and reference samples to build a writer-dependent classifier. After extracting interest points from both images, they generated a set of 12 features, using information such as the number of SIFT matches between the two images, and processing time. 
Malik et al. \cite{malik_automatic_2014} used SURF to extract interest points in the signature images, and used these features to assess the local stability of the signatures. During classification, only the stable interest points are used for matching. The number of keypoints in the query image, and the number of matched keypoints were used to classify the signature as genuine or forgery.

\subsubsection{Pseudo-dynamic features}

Oliveira et al. \cite{oliveira_graphology_2005} presented a set of pseudo-dynamic features, based on graphometric studies: \emph{Distribution of pixels}, \emph{Progression} - that measures the tension in the strokes, providing information about the speed, continuity and uniformity, \emph{Slant} and \emph{Form} - measuring the concavities in the signature.

More recently, Bertolini et al. \cite{bertolini_reducing_2010} proposed a descriptor that considers the \emph{curvature} of the signature. This was accomplished by fitting Benzier curves to the signature outline (more specially, to the largest segment of the signature), and using the parameters of the curves as features.

\subsection{Deep learning}

There has been an increased interest in recent years on techniques that do not rely on hand-engineered feature extractors. Instead, feature representations are learned from \textit{raw} data (pixels, in the case of images). This is the case of Deep Learning models \cite{bengio_learning_2009}, \cite{lecun_deep_2015}. 

Early work applying representation learning for the task used private datasets and did not report much success: Ribeiro et al \cite{ribeiro_deep_2011} used RBMs to learn a representation for signatures, but only reported a visual representation of the learned weights, and not the results of using such features to discriminate between genuine signatures and forgeries. Khalajzadeh \cite{khalajzadeh_persian_2012} used CNNs for Persian signature verification, but only considered random forgeries in their tests.

Considering work that targeted the classification between genuine signatures and skilled forgeries, we find two main approaches in recent literature: 1) learning writer-independent features in a subset of users, to be used for training writer-dependent classifiers \cite{hafemann_ijcnn_2016, hafemann_icpr_2016, hafemann_pr_2017, zhang_multi-phase_2016}; 2) learning feature representations and a writer-independent system at once, using metric learning \cite{rantzsch_signature_2016}.

Hafemann et al. \cite{hafemann_ijcnn_2016} proposed a Writer-Independent feature learning method, where a development set $\mathcal{D}$ is used to learn a feature representation $\phi(X)$. This representation is learned using a Convolutional Neural Network (CNN) to discriminate among users in $\mathcal{D}$. After the network is trained, the function $\phi(X)$ is used as a feature extractor for the exploitation set $\mathcal{E}$, for which Writer-Dependent classifiers are trained. In later work \cite{hafemann_pr_2017}, the authors also proposed a multi-task framework, where the CNN is trained with both genuine signatures and skilled forgeries, optimizing to jointly discriminate between users, and discriminate between genuine signatures and forgeries.  Zhang et al. \cite{zhang_multi-phase_2016} proposed using Generative Adversarial Networks (GANs) \cite{goodfellow_generative_2014} for learning the features from a subset of users. In this case, two networks are trained: a generator, that learns to generates signatures, and a discriminator, that learns to discriminate if an image is from a real signature or one that was automatically generated. After training, the authors used the convolutional layers of the discriminator as the features for new signatures.

Rantzsch et al. \cite{rantzsch_signature_2016} proposed a Writer-Independent approach using metric learning. In this approach, the system learns a distance between signatures. During training, tuples composed of three signatures are fed to the network: ($X_r$, $X_+$, $X_-$), where $X_r$ is a reference signature, $X_+$ is a genuine signature from the same user, and $X_-$ is a forgery (either a random or skilled forgery). The system is trained to minimize the distance between $X_r$ and $X_+$, and maximize the distance between $X_r$ and $X_-$. The central idea is to a learn a feature representation that will therefore assign small distances when comparing a genuine signature to another (reference) genuine signature, and larger distances when comparing a skilled forgery with a reference.

\section{Model Training}
\label{sec:model_training}

As introduced in section \ref{sec:problem_statement}, the classifiers for signature verification can be broadly classified in two groups: \emph{writer-dependent} and \emph{writer-independent}. In the first case, which is more common in the literature, a model is trained for each user, using the user's genuine signatures, and random forgeries (by using genuine signature from other users). During the operations phase, the model trained for the claimed identity is used to classify query signatures as genuine or forgery. The writer-independent approach, on the other hand, involves only a single classifier for all users. In this case, the system learn to compare a query signature with a reference. During the test phase, the model is used to compare a query signature with reference genuine samples from the claimed individual, to make a decision. One common way of training WI systems is to use a dissimilarity representation, where the inputs to the classifiers are differences between two feature vectors: $|x_q - x_r|$, with a binary label that indicates whether the two signatures are from the same user or not \cite{rivard_multi-feature_2013, eskander_hybrid_2013}.

Some authors use a combination of both approaches. For example, Eskander et al. \cite{eskander_hybrid_2013} and Zhang et al. \cite{zhang_multi-phase_2016} trained hybrid writer-independent-writer-dependent solutions, where a writer-independent classifier is used for classification when only a few genuine signatures are available. When the number of collected genuine samples passes a threshold, a writer-dependent classifier is trained for the user. Yilmaz  \cite{yilmaz_score_2016} propose a hybrid approach, where the results of both a writer-independent and writer-dependent classifiers are combined. 

Besides the most basic classifiers (e.g. simple thresholding and nearest-neighbors), several strategies have been tried for the task of signature verification. The following sections cover the main models used for the task.

\subsection{Hidden Markov Models}

Several authors have proposed using Hidden Markov Models for the task of signature verification \cite{justino_off-line_2000}, \cite{oliveira_graphology_2005},  \cite{batista_dynamic_2012}. In particular, HMMs with a left-to-right topology have been mostly studied, as they match the dynamic  characteristics of American and European handwriting (with hand movements from left to right).

In the work from Justino \cite{justino_off-line_2000}, Oliveira \cite{oliveira_graphology_2005} and Batista \cite{batista_dynamic_2012}, the signatures are divided in a grid format. Each column of the grid is used as an observation of the HMM, and features are extracted from the different cells within each column, and subsequently quantized in a codebook. In the verification phase, a sequence of feature vectors is extracted from the signature and quantized using the codebook. The HMM is then used to calculate the likelihood of the observations given the model. After calculating the likelihood, a simple threshold can be used to discriminate between genuine signatures and forgeries \cite{justino_off-line_2000}, or the likelihood itself can be used for more complex classification mechanisms \cite{batista_dynamic_2012}.

\subsection{Support Vector Machines}

Support Vector Machines have been extensively used for signature verification, for both writer-dependent and writer-independent classification \cite{ozgunduz_off-line_2005}, \cite{justino_comparison_2005}, \cite{bertolini_reducing_2010}, \cite{kumar_writer-independent_2012}, \cite{yilmaz_score_2016}, \cite{hafemann_pr_2017}, empirically showing to be the one of the most effective classifiers for the task. In recent years, Guerbai et al  \cite{guerbai_effective_2015} used One-Class SVMs for the task. This type of model attempt to only model one class (in the case of signature verification, only the genuine signatures), which is a desirable property, since for the actual users enrolled in the system we only have the genuine signatures to train the model. However, the low number of genuine signatures present an important challenge for this strategy.

\subsection{Neural Networks and Deep Learning}

Neural Networks have been explored for both writer-dependent and writer-independent systems. Huang and Yan \cite{huang_off-line_1997} used Neural Networks to classify between genuine signatures and random and targeted forgeries. They trained multiple networks on features extracted at different resolutions, and another network to make a decision, based on the outputs of these networks. Shekar et al \cite{shekar_grid_2015} presented a comparison of neural networks and support vector machines in three datasets.

More recently, Soleimani et al. \cite{soleimani_deep_2016} proposed a Deep Multitask Metric Learning (DMML) system for signature verification. In this approach, the system learns to compare two signatures, by learning a distance metric between them. The signatures are processed using a feedforward neural-network, where the bottom layers are shared among all users (i.e. the same weights are used), and the last layer is specific to each individual, and specializes for the individual.
In the work of Rantzsch et al. \cite{rantzsch_signature_2016}, a metric learning classifier is learned, jointly learning a feature representation, and a writer-independent classifier.

\subsection{Ensemble of classifiers}

Some authors have adopted strategies to train multiple classifiers, and combine their predictions when classifying a new sample. Bertollini et al. \cite{bertolini_reducing_2010} used a static ensemble selection with graphometric features. They generate a large pool of classifiers (trained with different grid sizes), and used a genetic algorithm to select a subset of the models, building an ensemble of classifiers.  
Batista et al \cite{batista_dynamic_2012} used dynamic selection of classifiers for building a writer-dependent system. They used a bank of HMMs as base classifiers, and for a given sample, the posterior likelihood is calculated for all HMMs. The set of likelihoods is considered as a feature vector, and a specialized random subspace method is used to train an ensemble of classifiers . 
Yilmaz and Yanikoglu \cite{yilmaz_score_2016} proposed a system that combines writer-dependent and writer-independent models (trained with a variety of feature descriptors). The scores from all classifiers is subsequently aggregated using a linear combination, obtaining a final decision of the ensemble.

%

\subsection{Data augmentation}

One of the main challenges for building an automated signature verification system is the low number of samples per user for training. To address this issue, some researchers have proposed ways to generate more samples based on existing genuine signatures. 

Huang and Yan \cite{huang_off-line_1997} have proposed a set of ``perturbations" to be applied to each genuine  signature, to generate new samples: slant, rotation, scaling and perspective. In their work, they considered a set of ``slight distortions", used to create new genuine samples, and ``heavy distortions" to generate forgeries from the genuine samples. More recently, Ferrer et al \cite{ferrer_synthetic_2013}, \cite{ferrer_static_2015}, \cite{diaz_generation_2016} have proposed a signature synthesis approach inspired on a neuromotor model.

\subsection{Classification performance on commonly used datasets}

\begin{table}
\centering
\caption{State-of-the-art performance on the GPDS dataset}

\resizebox{\columnwidth}{!}{%
\begin{tabular}{@{}llllcccc@{}} \toprule
  Type & Features and algorithm & \#Refs & FRR & FAR\textsubscript{skilled} &AER & EER\\ \midrule
  
WD \cite{vargas_off-line_2010} &Wavelets (SVM) &5 &  24.77&5.87&15.32&14.22\\
WD \cite{vargas_off-line_2011} &LBP, GLCM (SVM) &10 &  9.66&8.64&9.15&9.02\\
WD \cite{yilmaz_offline_2011} &LBP, HOG (SVM) &12 &  -&-&-&15.41\\
WD \cite{batista_dynamic_2012} &Pixel density (HMM + SVM) &12 &  16.81&16.88&16.85&-\\
WI \cite{kumar_writer-independent_2012} &Surroundness (NN) &1 &  -&-&-&13.76\\
WD \cite{bharathi_off-line_2013} &Chain Code (SVM) &12 &  13.16&9.64&11.4&-\\
WI \cite{eskander_hybrid_2013} &ESC + DPDF (Adaboost) &1 &  26.42&27.04&26.73&-\\
WD \cite{eskander_hybrid_2013} &ESC + DPDF (Adaboost) &14 &  18.06&22.71&20.39&-\\
WI \cite{hu_offline_2013} &LBP, GLCM, HOG (Adaboost) &1 &  -&-&-&9.94\\
WD \cite{hu_offline_2013} &LBP, GLCM, HOG (Adaboost) &10 &  -&-&-&7.66\\
WD \cite{guerbai_effective_2015} &Curvelet transform (OC-SVM) &12 &  12.5&19.4&15.95&-\\
WI \cite{yilmaz_score_2016} &LBP, HOG, SIFT (SVMs) &1 &  -&-&-&17.14\\
WD \cite{yilmaz_score_2016} &LBP, HOG, SIFT (SVMs) &12 &  -&-&-&6.97\\
WD \cite{shekar_grid_2015} & Pattern spectra (NN) &15 &  8.59&8.94&8.76&-\\
WD \cite{soleimani_deep_2016} &LBP (Metric Learning) &10 &  -&-&-&20.94\\
WD \cite{hafemann_pr_2017} & Feature learning (SVM) &12 &  3.94&3.53&3.73&1.69\\
WD \cite{zhang_multi-phase_2016} & \specialcell{Feature Learning \\(DCGAN + adaboost)} &14 &  -&-&\specialcell{12.57$\sim$\\16.08}&-\\

\bottomrule
\end{tabular}}
\label{table:soa_gpds}
\end{table}

\begin{table}
\centering
\caption{State-of-the-art performance on the MCYT dataset}

\resizebox{\columnwidth}{!}{%
\begin{tabular}{@{}llllcccc@{}} \toprule
  Type & Features and algorithm & \#Refs & FRR & FAR\textsubscript{skilled} &AER & EER\\ \midrule

WD \cite{gilperez_off-line_2008} &Contours (chi squared distance) &10 &  -&-&-&6.44\\
WD \cite{wen_model-based_2009} &RPF (HMM) &5 &  -&-&-&15.02\\
WD \cite{vargas_off-line_2011} &LBP (SVM) &10 &  8.69&6.54&7.62&7.08\\
WD \cite{ooi_image-based_2016} &DRT + PCA (PNN) &10 &  -&-&-&9.87\\
WD \cite{soleimani_deep_2016} &HOG (Metric Learning) &10 &  6.13&12.71&9.42&9.86\\
WD \cite{hafemann_pr_2017} & Feature learning (SVM) &10 &  -&-&-&2.87\\

\bottomrule
\end{tabular}}
\label{table:soa_mcyt}
\end{table}

\begin{table}
\centering
\caption{State-of-the-art performance on the CEDAR dataset}

\resizebox{\columnwidth}{!}{%
\begin{tabular}{@{}llllcccc@{}} \toprule
  Type & Features and algorithm & \#Refs & FRR & FAR\textsubscript{skilled} &AER & EER\\ \midrule

WD \cite{chen_new_2006} &Graph Matching &16 &  7.7&8.2&7.9&\\
WI \cite{kumar_writer-independent_2010} &Morphology (SVM) &1 &  12.39&11.23&11.81&11.59\\
WI \cite{kumar_writer-independent_2012} &Surroundness (NN) &1 &  &8.33&8.33&8.33\\
WD \cite{bharathi_off-line_2013} &Chain code (SVM) &12 &  9.36&7.84&7.84&\\
WD \cite{guerbai_effective_2015} &Curvelet transform (OC-SVM) &12 &  -&-&5.6&-\\
WD \cite{hafemann_pr_2017} &Feature learning (SVM) & 12 &  -&-&-&4.63\\

\bottomrule
\end{tabular}}
\label{table:soa_cedar}
\end{table}

Comparing the performances of different feature extractors and classifiers requires using standard datasets and similar (ideally equal) experimental protocols. The availability of public datasets enables the comparison of different approaches, although it must be noted that it is common to have differences in training protocol (e.g. number of references used for training, or how random forgeries are selected for training, among others) and testing protocol (e.g. which metrics to report). We consolidate the state-of-the-art performances on three datasets: GPDS in table \ref{table:soa_gpds}, MCYT in table \ref{table:soa_mcyt} and CEDAR in table  \ref{table:soa_cedar}, where the results are ordered by publication date. It is worth noting that the GPDS dataset had different releases (GPDS-100, GPDS-160, GPDS-300, GPDS-960) as more users were added to the dataset. In table \ref{table:soa_gpds} all results on the GPDS dataset have been grouped together.

For this comparison, we consider the type of the classifier (Writer-Dependent or Writer-Independent), the feature descriptors and classifiers, and the number of reference signatures used for training. For Writer-Independent systems, we report the number of samples required for the new users of the system (usually 1 reference signature). We consider the following metrics (when reported in the papers): False Rejection Rate (FRR) - the percentage of genuine signatures that are rejected by the system, False Acceptance Rate (FAR\textsubscript{skilled}) - the percentage of skilled forgeries that are accepted, Average Error Rate (AER): the average error considering only FRR and FAR\textsubscript{skilled}, and the Equal Error Rate (EER) - the error obtained when FAR = FAR\textsubscript{skilled}.

In general, we see that many feature extractors have been evaluated for the task, with texture descriptors (LBP and GLCM) and directional-based descriptors (HOG and DPDF) being particularly important. More recently, feature learning methods showed potential to the task, achieving the best performance on these datasets.

\section{Conclusion}

Over the last decade, researchers have proposed a large variety of methods for Offline Signature Verification. While distinguishing genuine signatures and skilled forgeries remains a challenging task, error rates have dropped significantly in the last few years, mostly due to advancements in Deep Learning applied to the task. 
Analyzing the recent contributions to the field, we can notice that they concentrate in the following categories:
\begin{itemize}
\item \textbf{Obtaining better features} - Several new feature extractors have been proposed for the task. Texture features (LBP variations), interest-point matching (SIFT, SURF) and directional features (HOG) have been successfully used to increase the accuracy of Offline Signature Verification Systems. More recently, feature learning methods have been successfully applied for the task, showing that features learned for a subset of users generalize to new users, and even users from other datasets.
\item \textbf{Improving classification with limited number of samples} - Given the severe constraints in practical applications, researchers have searched for ways to increase performance in cases where a small number of samples per user is available. In particular, the creation of dissimilarity-based writer-independent solutions, and metric-learning solutions have shown to be promising to address this problem.
\item \textbf{Augmenting the datasets} - Related to the problem of having low number of samples per user, some researchers have focused in generating synthetic signatures, in order to increase the number of samples available for training.
\item \textbf{Building model ensembles} - In order to increase classification accuracy, and the robustness of the solutions, some researchers have investigated the creation of both static and dynamic ensembles of classifiers.
\end{itemize}

In the authors' opinion, this trend will continue for future work, with researchers continuing to explore better feature representations (in particular learning representations from signature images with Deep Learning methods), and ways to improve classification with limited number of samples. Methods based on ensembles of classifiers, in particular techniques for dynamic selection \cite{britto_dynamic_2014,cruz_dynamic_2018} are also promising directions. Another problem that has not been sufficiently addressed in the literature is the usage of one-class classification models. One-class classifiers are theoretically interesting for this task, since they better match the problem statement. One-class classification systems that work well with low number of samples per user is an interesting area for future research.

\section*{Acknowledgment}

This work was supported by the Fonds de recherche du Qu\'ebec - Nature et technologies (FRQNT), the CNPq grant \#206318/2014-6 and by the grant RGPIN-2015-04490 to Robert Sabourin from the NSERC of Canada.

\bibliographystyle{plain}
\bibliography{report}

\end{document}